\documentclass[letterpaper, 10 pt, conference]{ieeeconf}
\let\labelindent\relax
\IEEEoverridecommandlockouts
\usepackage{cite}
\usepackage{amsmath,amssymb,amsfonts}
\usepackage{algorithmic}
\usepackage{graphicx}
\usepackage{textcomp}
\usepackage{xcolor}

\usepackage{multicol}
\usepackage{multirow}
\usepackage{adjustbox}
\usepackage{pgfplots}
    \pgfplotsset{compat=1.17}
\usepackage{tikz}
    \usetikzlibrary{shapes,arrows}
\usepackage{support-caption}
\usepackage[font={footnotesize}]{subcaption}
\usepackage[font={footnotesize}]{caption}
\usepackage[inline]{enumitem}
\usepackage{float}
\usepackage{balance}

\usepackage{siunitx}
\begin{document}

\title{ \LARGE \bf Real-time Geo-localization Using Satellite Imagery and \\Topography for Unmanned Aerial Vehicles}

\author{Shuxiao Chen, Xiangyu Wu, Mark W.\ Mueller and Koushil Sreenath}

\maketitle

\begin{abstract}
The capabilities of autonomous flight with unmanned aerial vehicles (UAVs) have significantly increased in recent times. However, basic problems such as fast and robust geo-localization in GPS-denied environments still remain unsolved. Existing research has primarily concentrated on improving the accuracy of localization at the cost of long and varying computation time in various situations, which often necessitates the use of powerful ground station machines. In order to make image-based geo-localization online and pragmatic for lightweight embedded systems on UAVs, we propose a framework that is reliable in changing scenes, flexible about computing resource allocation and adaptable to common camera placements. The framework is comprised of two stages: offline database preparation and online inference. At the first stage, color images and depth maps are rendered as seen from potential vehicle poses quantized over the satellite and topography maps of anticipated flying areas. A database is then populated with the global and local descriptors of the rendered images. At the second stage, for each captured real-world query image, top global matches are retrieved from the database and the vehicle pose is further refined via local descriptor matching. We present field experiments of image-based localization on two different UAV platforms to validate our results.
\end{abstract}

\section{Introduction}
Geo-localization for UAVs operating in GPS-denied environments has been an active research area due to its critical need for practical applications. This is because GPS signal loss is inevitable due to various unpredictable factors in real-world flight missions. Image-based geo-localization methods are intuitive and also biologically inspired. 
For an agent that is isolated from global localization sources, visual information is a reliable observation that the agent can actively collect for localization. 
As a growing number of satellite imagery sources are made public, they have unleashed the potential of image-based methods. Particularly, the more accessible topographic information of terrains and buildings offers a unique advantage.

An ideal set of capabilities for an image-based geo-localizer may contain the following aspects. 
%
Firstly, while multiple sensor inputs can be fused for localization, it should be able to localize purely with images. 
This requirement enhances the safety of the agent in the event of sensor failures. Next, it is desirable for the geo-localizer to use query images which are not captured strictly perpendicular or nadir to the ground, because the camera may not only be utilized for geo-localization but also have other purposes requiring capturing a forward view, e.g.\ obstacle avoidance at low elevations. This demand necessitates the localization outputs in more degrees-of-freedom (DOFs) as opposed to solely latitude and longitude. Furthermore, the geo-localizer should be ideally capable of being bootstrapped. In situations where no initial positions are available for the agent or long periods of GPS interruption is present, the geo-localizer needs to determine the position without prior knowledge. Ultimately, it is important to have the ability to infer location in real-time on low-power and low-compute UAV platforms.

\begin{figure}
    \centering
    \includegraphics[width=\linewidth]{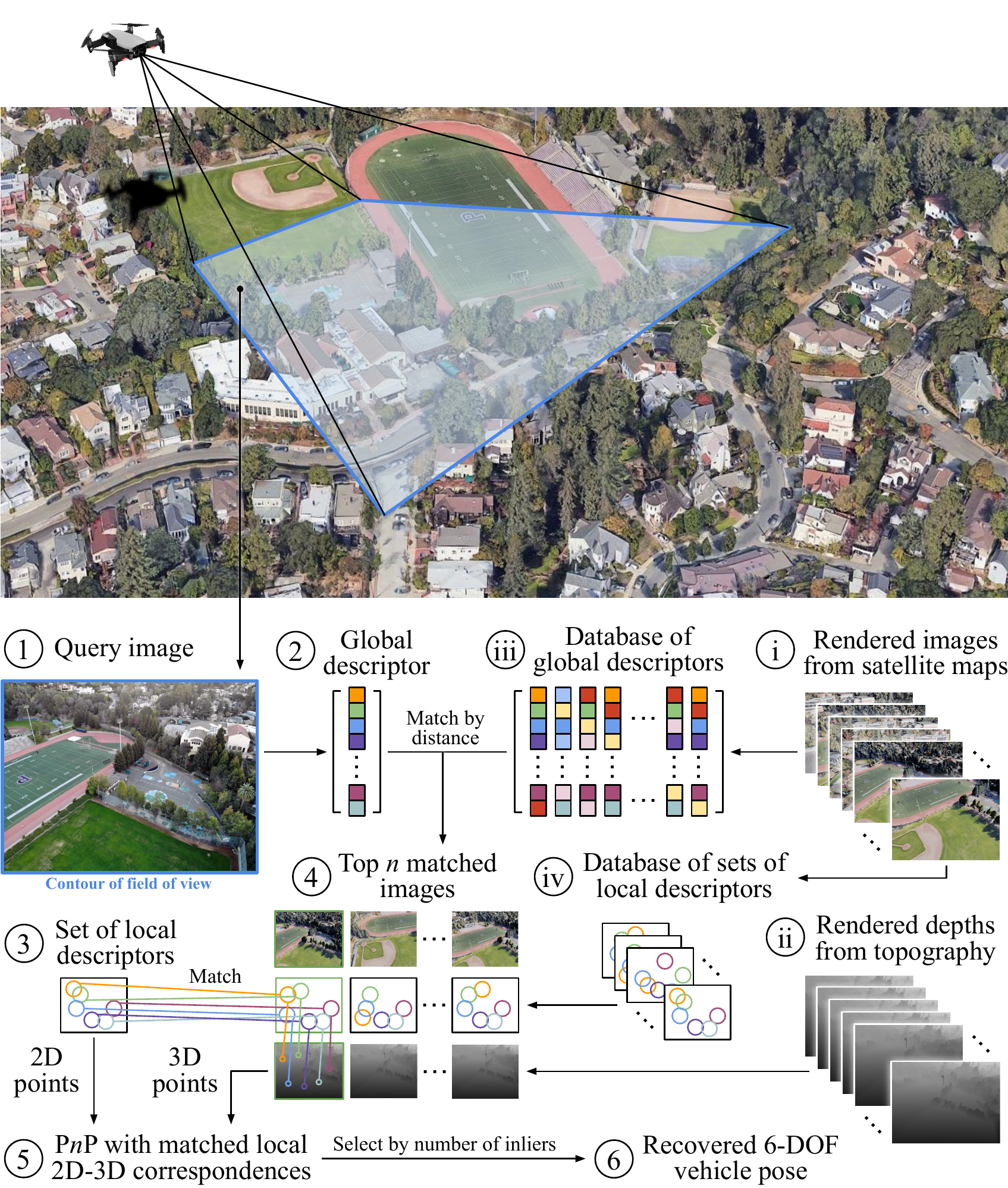}
    \caption{The proposed geo-localization pipeline. The steps numbered in Roman numerals (i-iv) introduce the procedure of the offline database preparation (Sec.\ \ref{sec:database-prep}) of the descriptor database and those in Arabic numerals (1-6) define the onboard inference (Sec.\ \ref{sec:onboard-inference}) based localization strategy. The pose and field of view of the quadrotor along with a captured query image from one of our experiments are shown.}
    \vspace{-3.0ex}
    \label{cover_pic}
\end{figure}

\begin{table*}
\caption{A comparison among popular geo-localization methods regarding a variety of attributes. Within each attribute, the methods which are regarded to produce results that are the most advantageous to aerial applications are highlighted in bold and colored red. The statistics are selected to demonstrate the nominal performance of each method by excluding the results from extreme cases presented in the corresponding papers.}
\vspace{-1ex}
\centering
\resizebox{\linewidth}{!}{%
\begin{tabular}{lcccccccc}
\hline
\multicolumn{1}{|c|}{\multirow{2}{*}{\textbf{Attribute}}} &
  \multicolumn{4}{c|}{\textbf{Direct alignment}} &
  \multicolumn{4}{c|}{\textbf{Feature matching}} \\ \cline{2-9} 
\multicolumn{1}{|c|}{} &
  \multicolumn{1}{c|}{NID \cite{patel2020visual}} &
  \multicolumn{1}{c|}{MI \cite{yol2014vision}} &
  \multicolumn{1}{c|}{ICLK with depth \cite{hinzmann2020deep}} &
  \multicolumn{1}{c|}{ICLK with homography \cite{goforth2019gps}} &
  \multicolumn{1}{c|}{Siamese NN \cite{shetty2019uav}} &
  \multicolumn{1}{c|}{SIFT \cite{wu2008image}} &
  \multicolumn{1}{c|}{Autoencoder \cite{bianchi2021uav}} &
  \multicolumn{1}{c|}{Ours} \\ \hline
\multicolumn{1}{|l|}{Stand-alone$^{\mathrm{a}}$} &
  \multicolumn{1}{c|}{No, needs VO} &
  \multicolumn{1}{c|}{No, needs IMUs} &
  \multicolumn{1}{c|}{\textcolor{red}{\textbf{Yes}}} &
  \multicolumn{1}{c|}{\textcolor{red}{\textbf{Yes}}} &
  \multicolumn{1}{c|}{No, needs VO} &
  \multicolumn{1}{c|}{\textcolor{red}{\textbf{Yes}}} &
  \multicolumn{1}{c|}{\textcolor{red}{\textbf{Yes}}} &
  \multicolumn{1}{c|}{\textcolor{red}{\textbf{Yes}}} \\ \hline
\multicolumn{1}{|l|}{Nadir camera} &
  \multicolumn{1}{c|}{Yes} &
  \multicolumn{1}{c|}{Yes} &
  \multicolumn{1}{c|}{\textcolor{red}{\textbf{No}}} &
  \multicolumn{1}{c|}{Yes} &
  \multicolumn{1}{c|}{\textcolor{red}{\textbf{No}}} &
  \multicolumn{1}{c|}{\textcolor{red}{\textbf{No}}} &
  \multicolumn{1}{c|}{Yes} &
  \multicolumn{1}{c|}{\textcolor{red}{\textbf{No}}} \\ \hline
\multicolumn{1}{|l|}{Min. working height} &
  \multicolumn{1}{c|}{$36$-\SI{48}{\metre}} &
  \multicolumn{1}{c|}{$125$-\SI{175}{\metre}} &
  \multicolumn{1}{c|}{-} &
  \multicolumn{1}{c|}{$200$-\SI{220}{\metre}} &
  \multicolumn{1}{c|}{$100$-\SI{200}{\metre}} &
  \multicolumn{1}{c|}{-} &
  \multicolumn{1}{c|}{\SI{40}{\metre}} &
  \multicolumn{1}{c|}{\textcolor{red}{\textbf{\SI{18}{\metre}}}} \\ \hline
\multicolumn{1}{|l|}{Desired path} &
  \multicolumn{1}{c|}{Yes} &
  \multicolumn{1}{c|}{\textcolor{red}{\textbf{No}}} &
  \multicolumn{1}{c|}{\textcolor{red}{\textbf{No}}} &
  \multicolumn{1}{c|}{\textcolor{red}{\textbf{No}}} &
  \multicolumn{1}{c|}{\textcolor{red}{\textbf{No}}} &
  \multicolumn{1}{c|}{\textcolor{red}{\textbf{No}}} &
  \multicolumn{1}{c|}{Yes} &
  \multicolumn{1}{c|}{\textcolor{red}{\textbf{No}}} \\ \hline
\multicolumn{1}{|l|}{Initial position} &
  \multicolumn{1}{c|}{Yes} &
  \multicolumn{1}{c|}{Yes} &
  \multicolumn{1}{c|}{Yes} &
  \multicolumn{1}{c|}{Yes} &
  \multicolumn{1}{c|}{\textcolor{red}{\textbf{No}}} &
  \multicolumn{1}{c|}{\textcolor{red}{\textbf{No}}} &
  \multicolumn{1}{c|}{Yes} &
  \multicolumn{1}{c|}{\textcolor{red}{\textbf{No}}} \\ \hline
\multicolumn{1}{|l|}{Degrees of freedom$^{\mathrm{b}}$} &
  \multicolumn{1}{c|}{4} &
  \multicolumn{1}{c|}{4} &
  \multicolumn{1}{c|}{\textcolor{red}{\textbf{6}}} &
  \multicolumn{1}{c|}{\textcolor{red}{\textbf{6}}} &
  \multicolumn{1}{c|}{\textcolor{red}{\textbf{6}}} &
  \multicolumn{1}{c|}{\textcolor{red}{\textbf{6}}} &
  \multicolumn{1}{c|}{3} &
  \multicolumn{1}{c|}{\textcolor{red}{\textbf{6}}} \\ \hline
\multicolumn{1}{|l|}{onboard$^{\mathrm{c}}$} &
  \multicolumn{1}{c|}{No} &
  \multicolumn{1}{c|}{No} &
  \multicolumn{1}{c|}{No} &
  \multicolumn{1}{c|}{No} &
  \multicolumn{1}{c|}{No} &
  \multicolumn{1}{c|}{No} &
  \multicolumn{1}{c|}{No} &
  \multicolumn{1}{c|}{\textcolor{red}{\textbf{Yes}}} \\ \hline
\multicolumn{1}{|l|}{Real-time$^{\mathrm{d}}$} &
  \multicolumn{1}{c|}{No} &
  \multicolumn{1}{c|}{No} &
  \multicolumn{1}{c|}{\textcolor{red}{\textbf{Yes}}} &
  \multicolumn{1}{c|}{No} &
  \multicolumn{1}{c|}{\textcolor{red}{\textbf{Yes}}} &
  \multicolumn{1}{c|}{No} &
  \multicolumn{1}{c|}{\textcolor{red}{\textbf{Yes}}} &
  \multicolumn{1}{c|}{\textcolor{red}{\textbf{Yes}}} \\ \hline
\multicolumn{1}{|l|}{Computing platform} &
  \multicolumn{1}{c|}{-} &
  \multicolumn{1}{c|}{Xeon \SI{2.8}{\giga\hertz}} &
  \multicolumn{1}{c|}{GeForce RTX 2080 Ti} &
  \multicolumn{1}{c|}{\textcolor{red}{\textbf{Core i5 \SI{2.9}{\giga\hertz}}}} &
  \multicolumn{1}{c|}{GeForce GTX 1050} &
  \multicolumn{1}{c|}{-} &
  \multicolumn{1}{c|}{Core i7, Quadro P2000} &
  \multicolumn{1}{c|}{\textcolor{red}{\textbf{Jetson AGX Xavier}}} \\ \hline
\multicolumn{1}{|l|}{Inference frequency} &
  \multicolumn{1}{c|}{$<$ \SI{0.2}{\hertz}} &
  \multicolumn{1}{c|}{-} &
  \multicolumn{1}{c|}{\SI{4.83}{\hertz}} &
  \multicolumn{1}{c|}{\SI{0.12}{\hertz}} &
  \multicolumn{1}{c|}{\SI{2.22}{\hertz}} &
  \multicolumn{1}{c|}{-} &
  \multicolumn{1}{c|}{\textcolor{red}{\textbf{\SI{9.09}{\hertz}}}} &
  \multicolumn{1}{c|}{\SI{1.12}{\hertz}} \\ \hline
\multicolumn{1}{|l|}{Translational error} &
  \multicolumn{1}{c|}{\textcolor{red}{\textbf{\textless\ \SI{3}{\metre} (3D, RMSE)}}} &
  \multicolumn{1}{c|}{\SI{12.76}{\metre} (3D, RMSE)} &
  \multicolumn{1}{c|}{\textcolor{red}{\textbf{\SI{4.37}{\metre} (3D, MAE)}}} &
  \multicolumn{1}{c|}{\SI{10.68}{\metre} (3D, MAE)} &
  \multicolumn{1}{c|}{\SI{36.0}{\metre} (3D, RMSE)} &
  \multicolumn{1}{c|}{-} &
  \multicolumn{1}{c|}{\textcolor{red}{\textbf{\SI{1.81}{\metre} (2D, RMSE)}}} &
  \multicolumn{1}{c|}{\textcolor{red}{\textbf{\SI{2.82}{\metre} (3D, RMSE)}}} \\ \hline
\multicolumn{1}{|l|}{Image dimensions$^{\mathrm{e}}$} &
  \multicolumn{1}{c|}{$560\times315$} &
  \multicolumn{1}{c|}{\textcolor{red}{\boldmath{$650\times500$}}} &
  \multicolumn{1}{c|}{\textcolor{red}{\boldmath{$752\times480$}}} &
  \multicolumn{1}{c|}{$200\times200$} &
  \multicolumn{1}{c|}{-} &
  \multicolumn{1}{c|}{-} &
  \multicolumn{1}{c|}{$320\times160$} &
  \multicolumn{1}{c|}{\textcolor{red}{\boldmath{$640\times480$}}} \\ \hline
\multicolumn{9}{l}{$^{\mathrm{a}}$Requires no additional sensors or estimate sources to assist the localization.\quad $^{\mathrm{b}}$The number of DOFs of the resulting poses directly from the geo-localizer.\quad $^{\mathrm{e}}$The dimensions are defined as width $\times$ height.} \\
\multicolumn{9}{l}{$^{\mathrm{c}}$Both data collection and pose estimation are finished onboard on the vehicle.\quad $^{\mathrm{d}}$The frequency of localization should not be too slow compared to the cruising speed and here \SI{1}{\hertz} is regarded as the boundary.}
\end{tabular}%
}
\vspace{-3.5ex}
\label{geoloc_performances_comparison}
\end{table*}

In order to meet the above requirements and address existing issues, we propose an image-based geo-localization pipeline shown in Fig.~\ref{cover_pic}. This pipeline consists of two phases. During the first offline database preparation phase, potential vehicle poses are sampled at a nominal flight attitude within a specified prospective flying area. Next, color images and depth maps are rendered as seen from the above sampled vehicle poses. Eventually, a database containing the global or whole-image descriptors and local descriptors of the rendered images, along with the depth maps, is constructed and loaded on the vehicle. The second phase is to perform online localization onboard the vehicle. This is achieved as follows. For each incoming query image collected by the UAV, the image is firstly encoded as a global descriptor and a set of local descriptors using neural networks (NNs). The encoded global descriptor is then matched over the database to find the most similar candidates. After that, the correspondences between the query and candidate sets of local descriptors are matched using NNs again. The absolute global coordinates of the matched correspondences are retrieved from the rendered depth maps. Ultimately, by combining the global coordinates and their projections on the query images, a Perspective-$n$-Point (P$n$P) problem is formulated and solved. We next summarize our contributions below.
\begin{itemize}
    \item We present a pragmatic image-based geo-localization package that executes on embedded systems at a high frequency.
    \item Our approach has the ability to adapt to downward-tilted camera configurations that are not rigorously nadir, therefore exploiting the existing setups.
    \item Our approach does not require a desired flight path for preparing the map database, thus enabling more flexible maneuvers.
    \item Our approach does not require prior knowledge of the initial positions for online localization, making it easier to choose the take-off location for the UAV.
\end{itemize}

This paper is structured as follows. It begins with background knowledge and related work in Sec.~\ref{sec:background} where image-based geo-localization, feature descriptors and camera pose estimation are covered respectively. After that, we introduce our proposed method in Sec.~\ref{sec:methodology} which is divided into the database preparation and onboard inference parts. The experimental setups and results are presented in Sec.~\ref{sec:experiment} along with the detailed procedure and analyses. Finally, we conclude with future work in Sec.~\ref{sec:conclusions}.

\section{Background and related work}
\label{sec:background}
We present the necessary background knowledge of image-based localization methods in this section. A brief survey of popular approaches can be found in TABLE \ref{geoloc_performances_comparison}.

\subsection{Image-based geo-localization}
In the early years of research into global localization of aerial vehicles using satellite maps, approaches largely concentrated on binarizing the query images and directly aligning them over a geo-referenced map database \cite{conte2008integrated}. Key assumptions were made in early work to simplify the problem, such as the low complexity in the scenes and high elevations at which the experiments were operated. After the emergence of key point detectors and feature descriptors which offer more compact representations of the useful information contained in an image, these were then employed in the geo-localization field as a new genre, named the feature matching method, e.g.\ the robust Scale-Invariant Feature Transform (SIFT) detector and descriptor \cite{lowe2004distinctive}. More detailed and advanced feature matching methods include matching the query images over an entire target map \cite{wu2008image} and individual mosaics of the map \cite{viswanathan2014vision} have been developed.

However, the handcrafted descriptors are computationally expensive and vulnerable to scene changes incurred by different seasons and lighting conditions. The first genre, the direct alignment method, again evolved to have the edge over the feature matching method in accuracy and scene tolerance. Mutual Information (MI) and optical flow are two typical modern direct alignment methods. They align two images in an incremental manner that does not require explicit presence of specific and similar key points in both images. Relative shift and rotation between images are parameterized to be the decision variables and difference metrics such as MI are minimized. Recent research shows that the method is capable of achieving superb localization accuracy whilst allowing inevitable discrepancies between query and database images, although approximate pose guesses are essential to initializing the optimization \cite{patel2020visual,hinzmann2020deep,yol2014vision,goforth2019gps}.

To eliminate the need for initial guesses, cross-view image-based hierarchical localization algorithms such as HF-Net \cite{sarlin2019coarse,sarlin2018leveraging}, or even those focusing on matching ground views to aerial imagery \cite{hu2020image,hu2018cvm,tian2017cross}, are inspiring for aerial geo-localization applications. The general philosophy that these algorithms follow is to invoke the power of NNs to semantically summarize the query and database images, either globally or locally, in order to tolerate the large visual difference caused by cross-view deformation and scene changes. Then, they formulate the localization problem as a retrieval task over a database consisting of mosaics of the collected maps. However, these databases are usually formed with real-world images which are not always available in aerial cases.

\subsection{Image and feature descriptors}
Feature descriptors play an important role in image retrieval problems. From the coverage perspective, descriptors are mainly categorized into global and local ones. Global descriptors, also known as whole-image descriptors, condense and encode the entire image to a lower dimensional space where similarity measures, or distance metrics, between images are easier to apply. Local descriptors aim at specific key points on the image and encode the content around the key points. Handcrafted descriptors, e.g.\ SIFT and Speeded Up Robust Features (SURF) \cite{bay2006surf}, etc., have been used as both whole-image \cite{viswanathan2014vision} and local descriptors \cite{wu2008image} for localization by adjusting their effective radii.

Because of the limited complexity and handcrafted nature of these descriptors, they were then replaced by NN-based methods such as NetVLAD \cite{arandjelovic2016netvlad} that aggregates local descriptors, which can be SIFT, SURF or even NN-based ones such as SuperPoint \cite{detone2018superpoint}, to encode the whole image.

\subsection{Pose estimation}
Localizing an agent in three-dimensional (3D) space with 6 DOFs is widely adopted in the simultaneous localization and mapping (SLAM) community because of the short relative distance between the agent and environment. Nevertheless, by imposing a flat-ground assumption due to imaging from higher elevations, most of the aforementioned geo-localization algorithms can then target more concise outputs with fewer DOFs by excluding the altitude, pitch and roll of the vehicle. This flat-ground assumption also validates the use of 2D satellite maps which are more easily available. On the other hand, geo-localization algorithms which do not require a rigorously nadir camera usually output full 6-DOF poses by decomposing the calculated planar homography to a set of rotations and translations \cite{goforth2019gps}.

With more frequently updated modern 3D map sources such as Google Earth, it has become possible to reduce the minimum working height and to compute and output poses of more DOFs by replacing the flat-ground assumption with richer depth information, thus allowing the geo-localization to be formulated as a P$n$P problem.

\section{Methodology}
\label{sec:methodology}
In this section, we present our methodology that is primarily divided into two parts: (1) the offline \emph{database preparation} given a potential flight venue and (2) the \emph{onboard inference} pipeline for real-time localization. Fig.~\ref{cover_pic} gives a full visual overview of the two parts.

\subsection{Database preparation}
\label{sec:database-prep}
The essence of image-based localization is the retrieval of recorded information at an acceptable speed and accuracy. Instead of storing the entire satellite map as a whole, cropped mosaics of the map are generated (see Sec.\ \ref{sec:image-rendering} on image rendering) and the mosaics are encoded into a lower dimensional space (see Sec.\ \ref{sec:image-encoding} on image encoding).

\subsubsection{Image rendering}
\label{sec:image-rendering}
Google Earth offers high-quality Earth maps in the sense of high-resolution satellite imagery, fine-grained 3D models of ground buildings and comprehensive topography information. We begin by retrieving these three parts from Google Earth via an open-source map builder \cite{retroplasma2021}. Next, we use Blender to stitch the downloaded colored meshes into a single piece according to their coordinates and rotate to fit the built-in $x$-$y$ plane. To achieve the coordinate switch, we override the latitude-longitude geographic coordinate system with the relative metric coordinate system. We call this the Blender coordinate system and define the origin arbitrarily to the stitched mesh. The geographic coordinate of the Blender origin is recorded. By assuming a negligible variation in the earth radius over the target terrain, the great-circle distance between a selected point and the Blender origin is regarded as the real-world distance. Conversely, to convert the distance back to the geographic coordinate, the Blender origin is used as the reference and the differences in distance are added or subtracted from it.

Next, we compute potential camera poses in the Blender coordinates in order to render images from the stitched map to generate mosaics. The ranges and spacings of the camera poses in $x$-direction, $y$-direction, elevation, heading and pitch angle are determined empirically in compliance with the area of the flight venue, the camera setup and the vehicle's onboard storage capacity. An effectual way of deciding the parameters for creating our database is detailed next.
\begin{enumerate}[label=(\roman*)]
    \item We begin with setting the simulated elevation to the nominal flight height of the mission. A camera with a vertical field of view (FOV) of around \SI{68}{\degree} flying at an elevation of about \SI{70}{\metre} can give $\pm29\%$ tolerance to the real vehicle height, i.e.\ with a single layer of potential camera poses simulated at an elevation of \SI{70}{\metre}, the real-world localization can then be achieved between around $50$-\SI{90}{\metre}. Additional layers of different elevations can be added to create hierarchical maps to increase the vertical working range of the vehicle.
    \item The spacings of the camera poses in the $x$- and $y$-directions are identical. In practice, this spacing is determined using the following two steps. Firstly, we obtain the rectangle projected by the camera when it is pointed strictly downward to a level ground. Secondly, a quarter of the shorter side of the rectangle is adopted as the spacing, e.g.\ for the aforementioned camera and nominal elevation, the spacing is \SI{10}{\metre}.
    \item A spacing of \SI{30}{\degree} in the heading is used for a camera with a horizontal FOV of \SI{84}{\degree}. It is recommended to have a resulting overlap of $50\%$ or above of the camera's horizontal FOV.
    \item The simulated pitch angle of the camera is aligned to the installation angle of the real camera. Images can be rendered with additional pitch angles for extra robustness of localization.
\end{enumerate}

Finally, the simulated images and corresponding depth maps are rendered at the spaced camera poses, as shown from Steps (i-ii) in Fig.\ \ref{cover_pic}. To maximize the amount of useful information contained in the 3D map, diffuse reflection is desired because any artificial shadows may cause unwanted visual discrepancies in the images. Hence, in Blender, the specular attribute in the material settings is tuned to the minimum possible value so that bright highlights on glossy surfaces are suppressed and the resulting reflections are more viewpoint-independent.

\subsubsection{Image encoding}
\label{sec:image-encoding}
Instead of storing the generated mosaic images of the map, we store the descriptors of the images.  There are a variety of choices of whole-image descriptors and compressed representation of images. NetVLAD \cite{arandjelovic2016netvlad} is employed in this case as the global descriptor because it is easily accessible in almost all deep learning frameworks. We use the version modified by \cite{sarlin2019coarse} to combine the features extracted by MobileNet \cite{sandler2018mobilenetv2} with a NetVLAD layer. Their pre-trained weights are used out-of-the-box to generate the global descriptors for the rendered images. For a simple representation in our implementation, each global descriptor is a $1$-by-$4096$ row vector and all global descriptors are vertically concatenated to form a $N$-by-$4096$ global descriptor array for $N$ rendered images. This aggregated global descriptor array is then serialized as a Python pickle binary file that is loaded onto the UAV.

SuperPoint \cite{detone2018superpoint} is applied as the local descriptor choice. Unlike handcrafted descriptors such as SIFT whose number of detected key points can vary, NN-based descriptors trigger a fixed amount of computation from end to end while also being more robust to scene changes. For each pose, the calculated local descriptors are serialized as a single file. The database containing these encoded global and local descriptors is presented as Steps (iii-iv) in Fig.\ \ref{cover_pic}.

Instead of matching descriptors via the brute-force matcher which iterates over all elements in the two sets without the awareness of semantic content, SuperGlue \cite{sarlin2020superglue} (proposed as a supporting matcher for SuperPoint) is used to find the correspondences between the local descriptors from two images. SuperGlue's pre-trained weights are utilized for inference due to its good exposure to outdoor street photos in the training set.

\subsection{Onboard inference}
\label{sec:onboard-inference}
After finishing the offline database preparation step, the 3D map and rendered images are discarded and only the serialized global descriptor array, the local descriptors and the depth maps are transferred to the aerial vehicle. NetVLAD, SuperPoint and SuperGlue network weights together with the calculated global descriptor array are de-serialized into the memory in advance for fast inference. When a query image is sent for inference, it is undistorted and down-sampled to match the dimensions of the prepared images, $640\times480$ in our case. We next detail the steps involved in online geo-localization.

\subsubsection{Matching strategy}
We begin by calculating a global descriptor for the query image captured from the UAV camera. The $\ell^2$-norm of the differences between the query image's global descriptor and every single global descriptor in the prepared array are computed and sorted in ascending order. After that, local descriptors are inferred for the query image and matched with the top $n$ candidates in the sorted list using SuperGlue. Here, $n$ is determined by the trade-off between performance and accuracy. Steps (1-4) in Fig.\ \ref{cover_pic} illustrate the procedure of the matching strategy.

\subsubsection{Vehicle pose recovery}
Using the matching strategy, within each pair of query and database global descriptors out of the $n$ matched pairs, the correspondences of local key points are found from their individual sets of local descriptors. In terms of each retrieved set of local descriptors from the database, the simulated depths of the correlated key points are obtained from its rendered depth map. Therefore, both the set of matched 3D coordinates on the map and their corresponding 2D projections in the query image are collected. OpenCV's built-in implementation of the P$n$P solver with RANSAC outlier rejection is used to recover the vehicle pose. Empirically, the re-projection error during the RANSAC procedure is set to 1.0 pixel and the number of iterations to 1000. Among the converged trials, the result from the one with the most inliers is treated as the final pose. A refinement threshold is set for the distance between the refined pose and its initially matched quantized position. If the pose correction from the refinement exceeds the threshold, i.e.\ the pose after P$n$P is too far away from the initially matched position, the trial is rejected. Eventually, the pose in the relative metric coordinate is converted to the geographic coordinate in latitude, longitude and altitude. Steps (5-6) in Fig.\ \ref{cover_pic} illustrate this pose recovery procedure.

\section{Experimental setups and results}
\label{sec:experiment}
Having shown our methodology for the geo-localization pipeline, we now introduce our experimental setups and present the field test results.

\begin{figure}
    \centering
    \includegraphics[width=.96\linewidth]{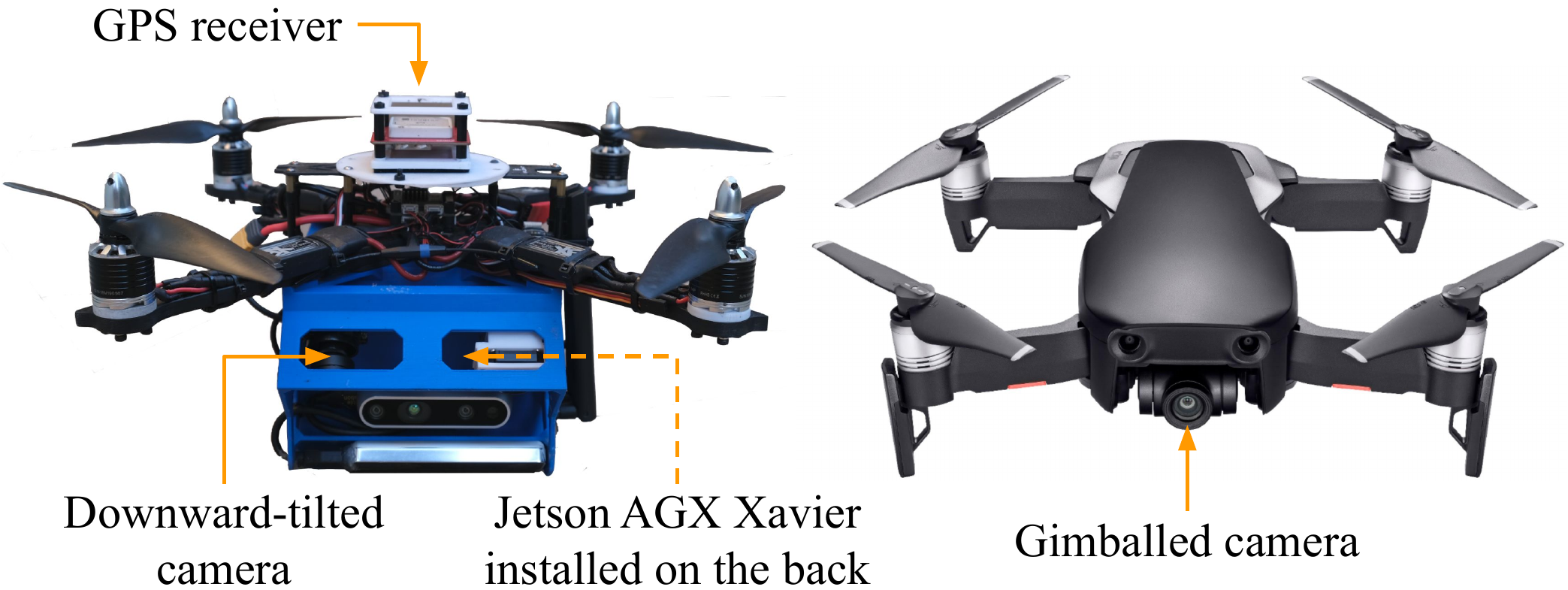}
    \vspace{-0.5ex}
    \caption{(Left) Front view of our customized quadrotor used at Location 1. Our camera is mounted and encapsulated by the blue cage. A GPS receiver is installed on the top to record ground truth trajectories. A Jetson AGX Xavier is attached to the back of the cage. (Right) The commercial DJI Mavic Air drone used to collect data at Location 2. The equipped gimballed camera can change the pitch angle resulting in a diversity of query images.}
    \vspace{-3.0ex}
    \label{customised_quadrotor}
\end{figure}

\subsection{Apparatus}
In order to test our pipeline's adaptability to images captured from various types of UAVs, we used two UAV platforms: a custom designed one as well as an inexpensive consumer UAV. As shown in Fig.\ \ref{customised_quadrotor}, our main vehicle was a customized quadrotor equipped with a passively cooled NVIDIA Jetson AGX Xavier as the primary computing resource. The Xavier possesses an 8-core Arch64 central processing unit (CPU) and a 512-core Volta graphics processing unit (GPU) to support the operating system and neural network inference. All motors and electronics were powered by a \SI{16.8}{\volt} lithium polymer battery with a capacity of \SI{5200}{\milli\ampere\hour}. The vision part consists of a global-shutter camera from IDS Imaging along with a wide-angle lens from Lensation. The resulting horizontal FOV of this camera package is \SI{84}{\degree} after image undistortion. We used an image aspect ratio of 4:3 resulting in dimensions of $640\times480$ pixels. This camera package was installed at \SI{45}{\degree} tilted downward from the horizon level. The camera was then able to cover the bottom-front view.

Our secondary platform is a commercial Mavic Air quadrotor from DJI. The Mavic Air's camera offers a horizontal FOV of around \SI{69}{\degree} and is electronically gimballed. We pitched down this gimballed camera to mimic similar viewing angles and down-sampled its images to the same dimensions of those from the customized quadrotor.

\subsection{Procedure}
Two experiments at two different locations were conducted. We aimed to verify the versatility of the pipeline by adding this variation in localization scenarios.

At test Location 1, our custom UAV was flown to collect images. At this location, an area of $\SI{400}{\metre}\times\SI{400}{\metre}$ was selected for building the database so that the potential vehicle poses were confined in this space. An interval of $\SI{10}{\metre}$ was used to quantize the camera poses along the latitude and longitude directions to generate the map mosaics. A fixed altitude of \SI{70}{\metre} was used as the nominal flight elevation. The yaw angles were spaced by \SI{30}{\degree} within one revolution thus $12$ directions in total. The pitch angle was fixed at \SI{45}{\degree} pointing to the bottom-front side. For these $19{,}200$ poses, blank areas without map information may appear in the rendered images if only the selected area of the map is used. This is because the tilted viewing angle sees more than the area that the camera FOV projects perpendicularly on the ground. Thus, in practice, a larger map of around $\SI{1}{\km}\times\SI{1.2}{\km}$ was employed to cover the actual visual range.

Regarding the run-time settings, at Location 1, the top $3$ matched candidates were used to enter the second stage for pose refinement via P$n$P. The refinement threshold was set to double the $x$- or $y$-direction spacing thus \SI{20}{\metre} in this case. All global and local descriptors were stored on the vehicle in the format of Python pickle. The depth maps were saved using floating-point OpenEXR format.

At test Location 2, the DJI Mavic Air was flown to collect images. The parameters for generating the database were adjusted according to the camera settings of the Mavic Air. During the flight, we varied the pitch angle from \SI{30}{\degree} to \SI{60}{\degree} instead of keeping it constant as at Location 1 so that the database images were rendered at pitch angles of \SI{30}{\degree}, \SI{45}{\degree} and \SI{60}{\degree} respectively. The nominal flight elevation was also reduced from \SI{70}{\metre} to \SI{60}{\metre}. Particularly, the horizontal FOV was narrowed from \SI{84}{\degree} to \SI{69}{\degree}. Query images were captured at \SI{0.5}{\hertz} and of the same dimensions as at Location 1 after down-sampling. The potential vehicle poses were limited within an area of $\SI{250}{\metre}\times\SI{250}{\metre}$. As this reduction in flying area releases more spare storage capacity for extra poses, the yaw angles were spaced by \SI{24}{\degree} thus $15$ directions in total. An identical $\SI{1}{\km}\times\SI{1.2}{\km}$ area was used to cover the visual range. The serialization of descriptors and depth maps was completed in the same manner as was done in the first experiment. $\mspace{0.5mu}$ The same refinement threshold was set. $\mspace{0.5mu}$ Since \begin{figure}[H]
    \centering
    \begin{subfigure}[t]{\linewidth}
        \centering
        \vspace{-0.4ex}
        \begin{tikzpicture}
    \begin{axis}[
            width=0.95\linewidth,
            font=\scriptsize,
            xlabel style={font=\footnotesize},
            ylabel style={font=\footnotesize},
            scale only axis,
            table/col sep=comma,
            enlargelimits=false,
            xmin=0,
            xmax=300,
            ymin=0,
            ymax=300,
            axis equal=true,
            axis equal image,
            xlabel={Distance in east-west direction [\si{\metre}]},
            ylabel={Distance in north-south direction [\si{\metre}]},
            legend style={row sep=-0.035cm}
        ]
        \addlegendimage{mark=x,only marks,red}
        \addlegendentry{Initial match}
        \addlegendimage{mark=*,yellow}
        \addlegendentry{Refined pose}
        \addlegendimage{orange}
        \addlegendentry{Correspondence}
        \addlegendimage{blue}
        \addlegendentry{Ground truth}
        \addplot graphics[xmin=-50,ymin=-50,xmax=350,ymax=350] {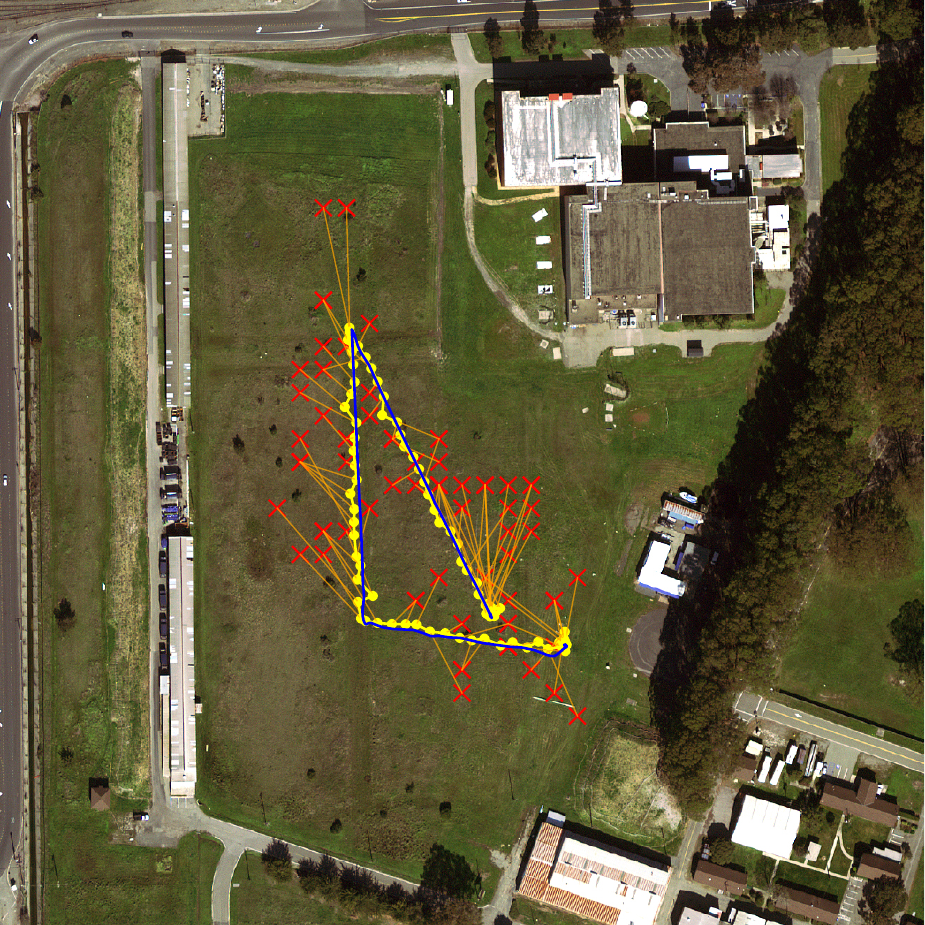};
    \end{axis}
\end{tikzpicture}
        \vspace{-3.7ex}
        \caption{The top view of the trajectory generated by our geo-localization pipeline is shown overlaid on a satellite map of Location 1. The red crosses are the positions of the initial matched candidates at which database images were generated. The yellow markers stand for the vehicle positions after the P$n$P refinement. The orange line segments show the correspondences between the initial matched positions and those after the refinement. The blue trajectory represents the ground truth positions from GPS. The length of the flight path is \SI{360}{\metre}.}
        \label{results_richmond_helicopter_view}
    \end{subfigure}
    \par\bigskip\smallskip
    \begin{subfigure}[t]{\linewidth}
        \centering
        \vspace{-2.2ex}
        \begin{tikzpicture}
    \begin{axis}[
            width=0.83\linewidth,
            height=0.23\linewidth,
            font=\scriptsize,
            xlabel style={font=\footnotesize},
            ylabel style={font=\footnotesize},
            scale only axis,
            table/col sep=comma,
            enlargelimits=false,
            xmin=-15,
            xmax=400,
            ymin=40,
            ymax=80,
            xlabel={Distance travelled projected on the ground [\si{\metre}]},
            ylabel={Altitude [\si{\metre}]},
            legend style={row sep=-0.035cm,at={(0.4,0.26)},anchor=east},
            ymajorgrids=true,
            ytick style={draw=none}
        ]
        \addlegendimage{blue,mark=x}
        \addlegendentry{Ground truth}
        \addlegendimage{red,mark=o}
        \addlegendentry{Inferred}
        \addplot[thick,blue,mark=x,mark options={scale=0.6},smooth] table[x=travel_dists,y=alt] {data/df_z_gt_richmond.csv};
        \addplot[thick,red,mark=o,mark options={scale=0.25},smooth] table[x=travel_dists,y=alt] {data/df_z_qr_richmond.csv};
    \end{axis}
\end{tikzpicture}
        \vspace{-3.7ex}
        \caption{A comparison between the inferred and ground truth altitudes versus the distance traveled by the vehicle. Although the vehicle was able to localize itself from \SI{45}{\metre} to \SI{73}{\metre} in elevation, all database images were generated at a single altitude of \SI{70}{\metre}.}
        \label{results_richmond_altitude_dist_travelled}
    \end{subfigure}
    \par\bigskip\smallskip
    \begin{subfigure}[t]{\linewidth}
        \centering
        \vspace{-1.3ex}
        \includegraphics[width=.24\textwidth]{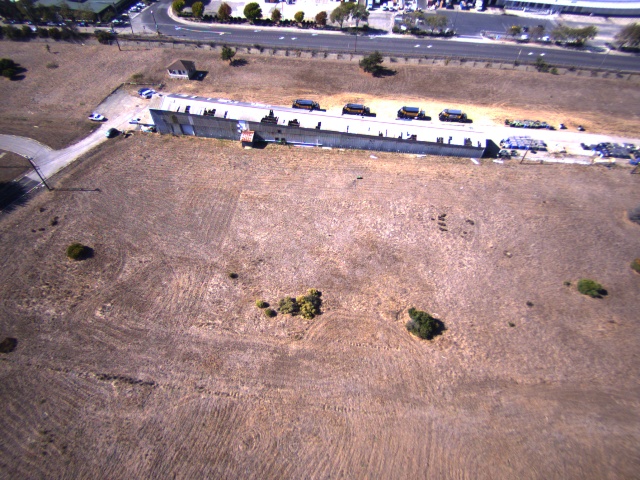}\hfill
        \includegraphics[width=.24\textwidth]{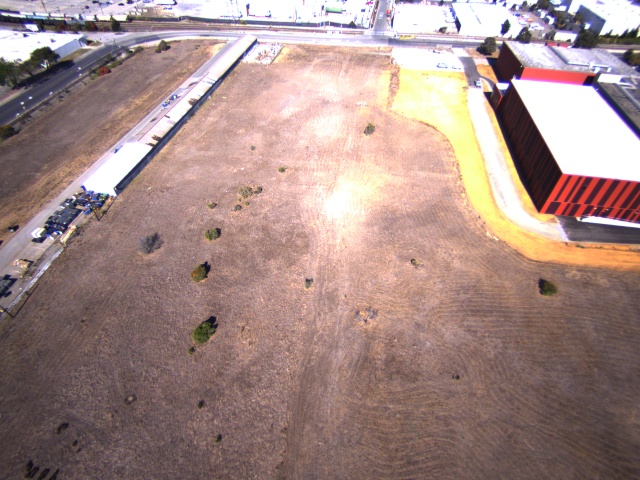}\hfill
        \includegraphics[width=.24\textwidth]{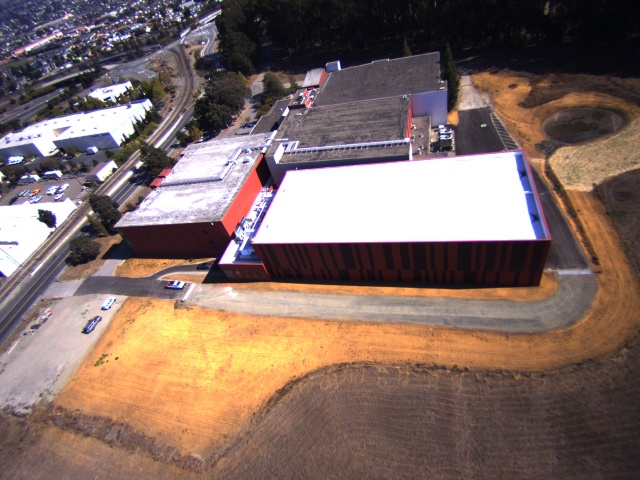}\hfill
        \includegraphics[width=.24\textwidth]{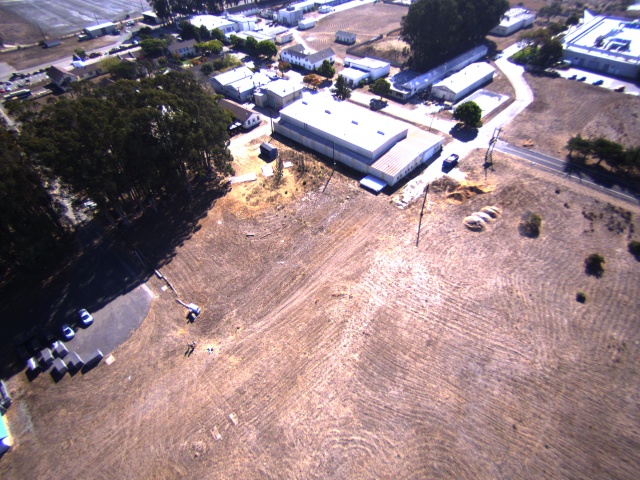}
        \\[\smallskipamount]
        \includegraphics[width=.24\textwidth]{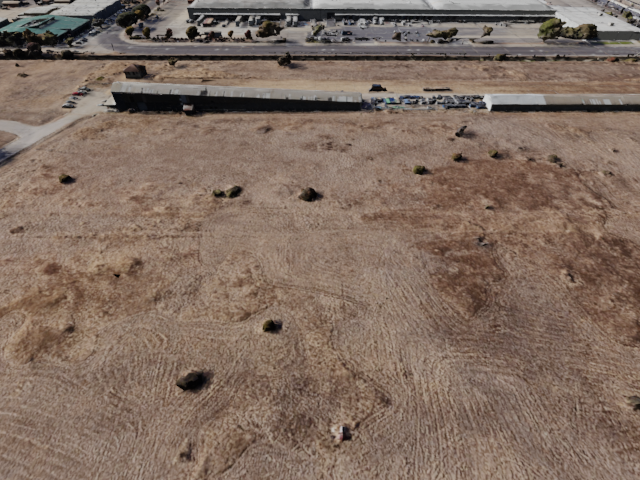}\hfill
        \includegraphics[width=.24\textwidth]{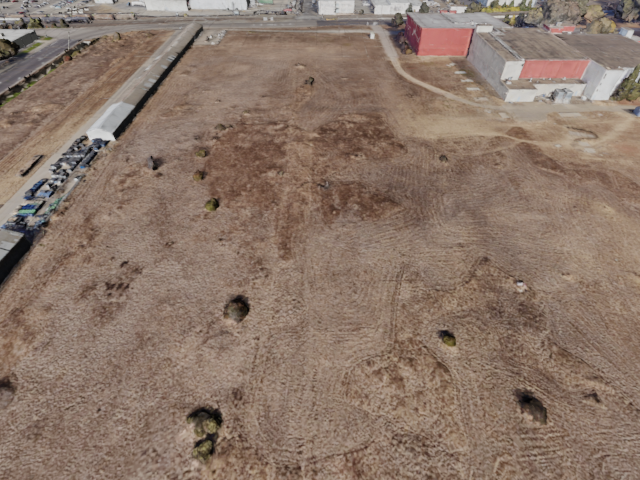}\hfill
        \includegraphics[width=.24\textwidth]{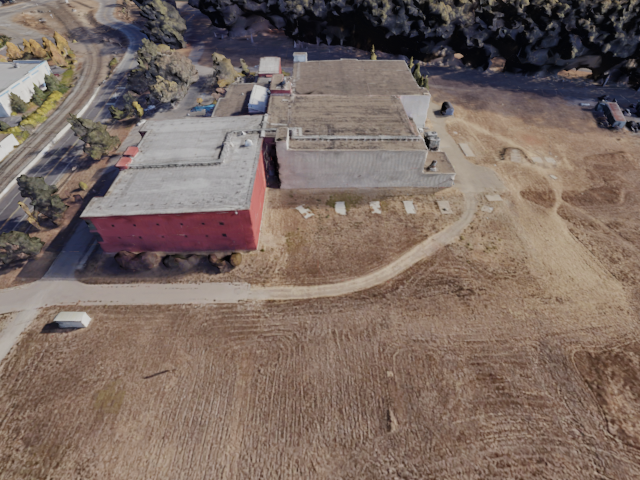}\hfill
        \includegraphics[width=.24\textwidth]{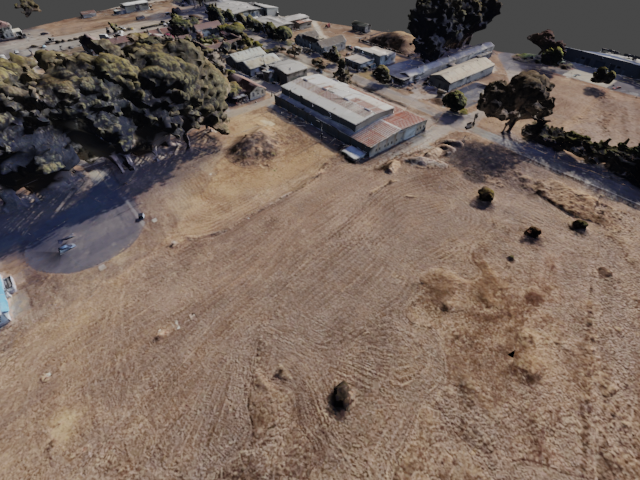}
        \caption{The top row shows the captured query images from the UAV and the bottom row presents the corresponding matched database images. In the two columns in the middle, there are significant discrepancies between the current query images and the dated satellite maps, e.g.\ the missing buildings in the scene.}
        \label{results_richmond_real_sim}
    \end{subfigure}
    \vspace{-0.8ex}
    \caption{Geo-localization results from the test flight at Location 1. Database was generated using satellite imagery from 2018 while query images were collected from experiments in 2021.}
    \vspace{-1.5ex}
    \label{results_richmond}
\end{figure}
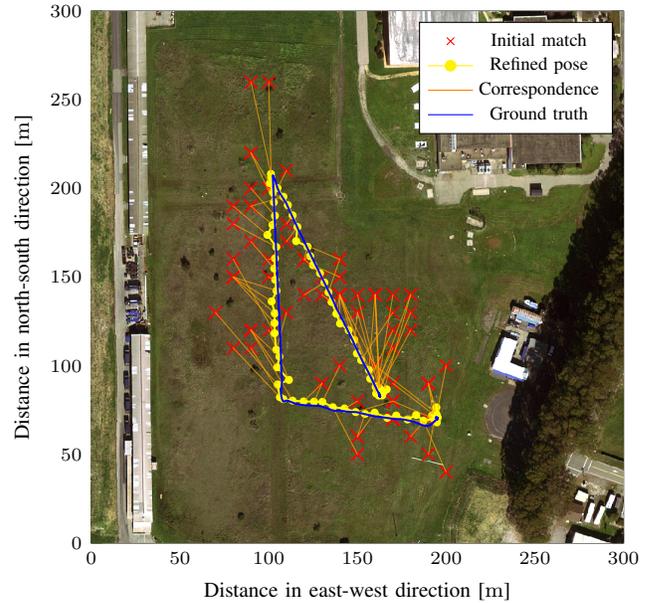
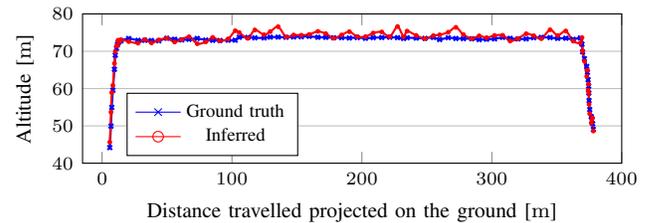
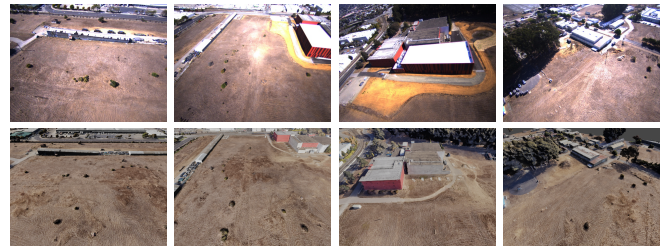
\noindent online inference was not possible on the Mavic Air due to the proprietary computation system, we ran the inference offline on a ground station. We still tested with different numbers of initial candidates that would enter the second stage for pose refinement via P$n$P.

\subsection{Results}
The visualization of the results from the first experiment conducted at Location 1 with three initial candidates is illustrated in Fig.\ \ref{results_richmond}. In this experiment, we reserved the top 3 matched global descriptors from the database and let them enter the pose refinement stage. The inferred and ground truth trajectories are overlaid on a satellite map and are demonstrated in Fig.\ \ref{results_richmond_helicopter_view}. The orientation of the satellite map in the background is aligned to the $y$-axis whose positive direction is strictly pointed to the North. In Fig.\ \ref{results_richmond_altitude_dist_travelled}, the inferred and ground truth altitudes are compared to show the performance of height estimation which is normally omitted in geo-localization benchmarks. The onboard localization pipeline was functioning during both the take-off and landing phases. The minimum localized altitude was \SI{45}{\metre} even with database images that were rendered at a nominal altitude of \SI{70}{\metre}. Fig.\ \ref{results_richmond_real_sim} shows a comparison between the real query images and database images. The localizer was able to overcome the significant discrepancies between the real scenes and outdated satellite maps with missing buildings.

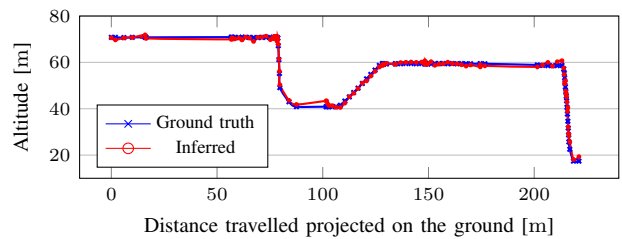
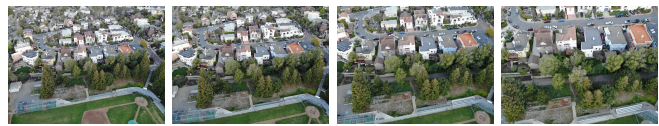
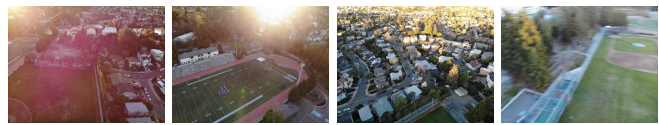
\begin{figure}[H]
    \centering
    \begin{subfigure}[t]{\linewidth}
        \centering
        \vspace{0.7ex}
        \begin{tikzpicture}
    \begin{axis}[
            width=0.95\linewidth,
            font=\scriptsize,
            xlabel style={font=\footnotesize},
            ylabel style={font=\footnotesize},
            scale only axis,
            table/col sep=comma,
            enlargelimits=false,
            xmin=0,
            xmax=300,
            ymin=0,
            ymax=300,
            axis equal=true,
            axis equal image,
            xlabel={Distance in east-west direction [\si{\metre}]},
            ylabel={Distance in north-south direction [\si{\metre}]},
            legend style={row sep=-0.035cm}
        ]
        \addlegendimage{mark=x,only marks,red}
        \addlegendentry{Initial match}
        \addlegendimage{mark=*,yellow}
        \addlegendentry{Refined pose}
        \addlegendimage{orange}
        \addlegendentry{Correspondence}
        \addlegendimage{blue}
        \addlegendentry{Ground truth}
        \addplot graphics[xmin=-20,ymin=-40,xmax=380,ymax=360] {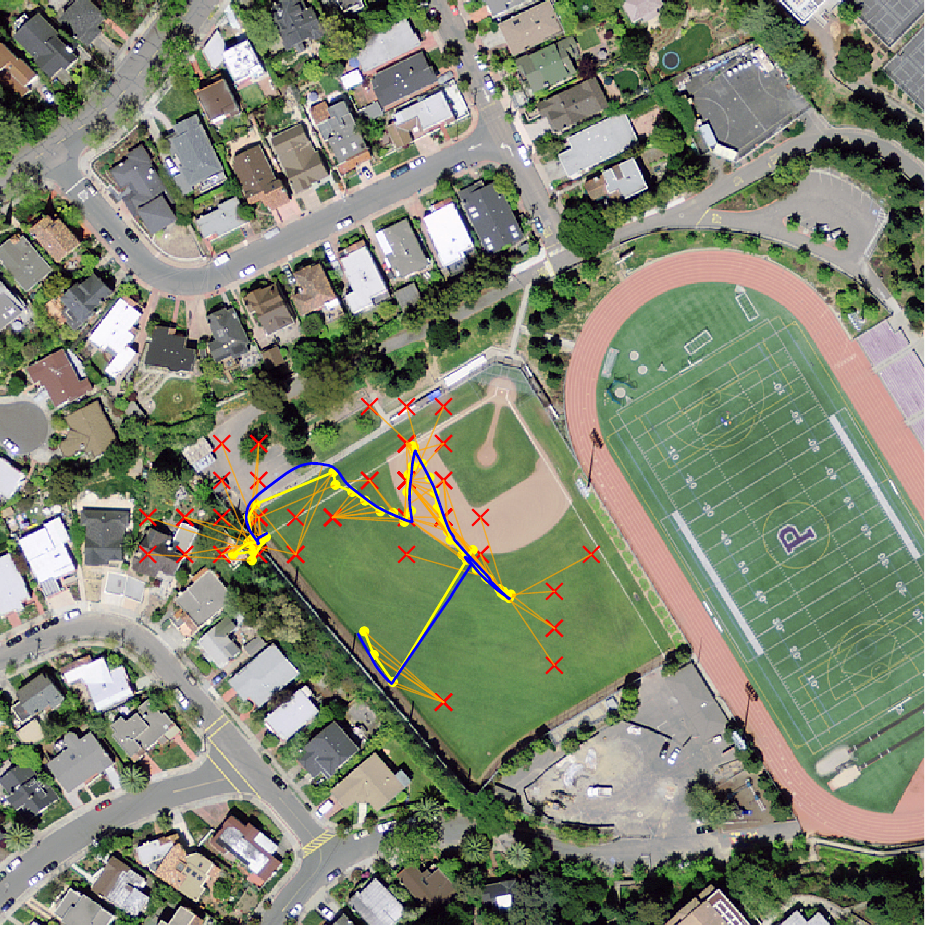};
    \end{axis}
\end{tikzpicture}
        \vspace{-3.6ex}
        \caption{Top view of inferred and ground truth trajectories overlaid on a map of Location 2. The visualization format shares the same fashion as in Fig.\ \ref{results_richmond_helicopter_view}. The length of the flight path is \SI{220}{\metre}.}
        \label{results_piedmont_helicopter_view}
    \end{subfigure}
    \par\bigskip\smallskip
    \begin{subfigure}[t]{\linewidth}
        \centering
        \vspace{-1.3ex}
        \begin{tikzpicture}
    \begin{axis}[
            width=0.83\linewidth,
            height=0.25\linewidth,
            font=\scriptsize,
            xlabel style={font=\footnotesize},
            ylabel style={font=\footnotesize},
            scale only axis,
            table/col sep=comma,
            enlargelimits=false,
            xmin=-15,
            xmax=240,
            ymin=10,
            ymax=80,
            xlabel={Distance travelled projected on the ground [\si{\metre}]},
            ylabel={Altitude [\si{\metre}]},
            legend style={row sep=-0.035cm,at={(0.35,0.26)},anchor=east},
            ymajorgrids=true,
            ytick style={draw=none}
        ]
        \addlegendimage{blue,mark=x}
        \addlegendentry{Ground truth}
        \addlegendimage{red,mark=o}
        \addlegendentry{Inferred}
        \addplot[thick,blue,mark=x,mark options={scale=0.6},smooth] table[x=travel_dists,y=alt] {data/df_z_gt_piedmont.csv};
        \addplot[thick,red,mark=o,mark options={scale=0.25},smooth] table[x=travel_dists,y=alt] {data/df_z_qr_piedmont.csv};
    \end{axis}
\end{tikzpicture}
        \vspace{-3.6ex}
        \caption{A comparison between the inferred and ground truth altitudes versus the distance traveled by the vehicle. Query images captured from \SI{18}{\metre} to \SI{70}{\metre} in elevation were successfully localized despite all database images being generated at a single altitude of \SI{60}{\metre}.}
        \label{results_piedmont_altitude_dist_travelled}
    \end{subfigure}
    \par\bigskip\smallskip
    \begin{subfigure}[t]{\linewidth}
        \centering
        \vspace{-0.8ex}
        \includegraphics[width=.24\textwidth]{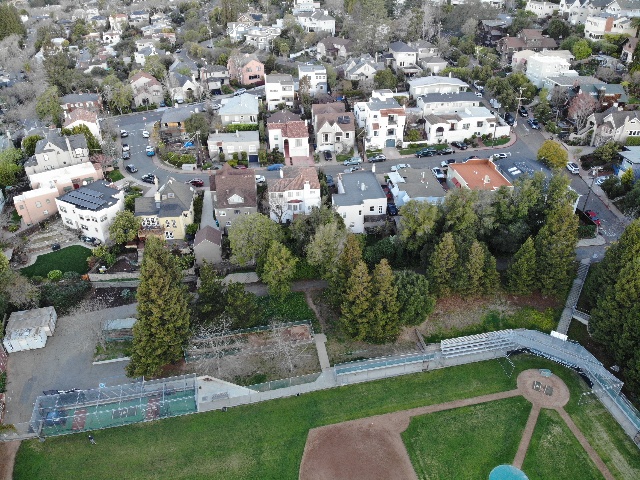}\hfill
        \includegraphics[width=.24\textwidth]{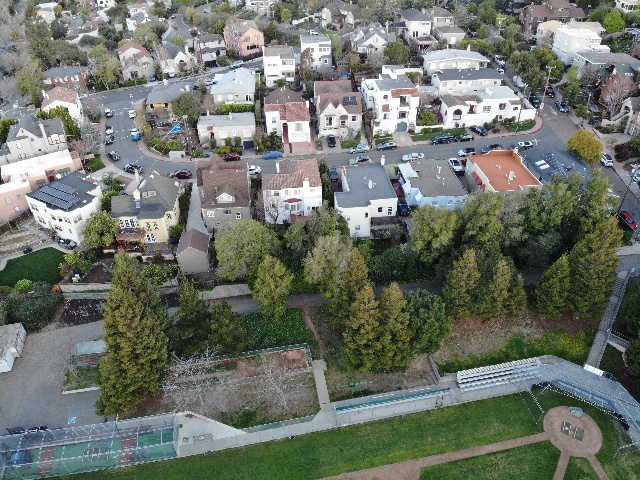}\hfill
        \includegraphics[width=.24\textwidth]{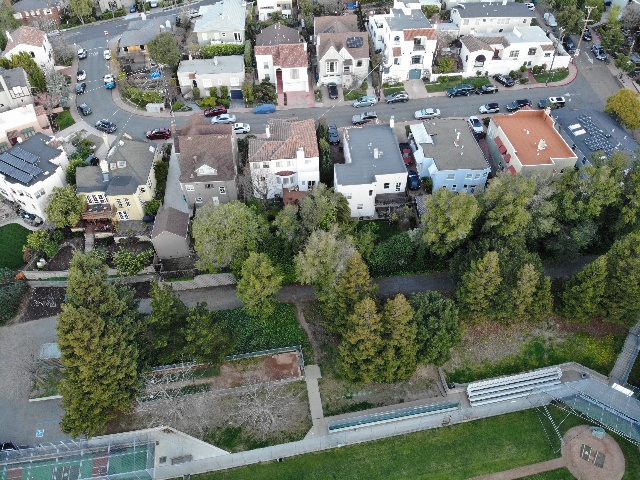}\hfill
        \includegraphics[width=.24\textwidth]{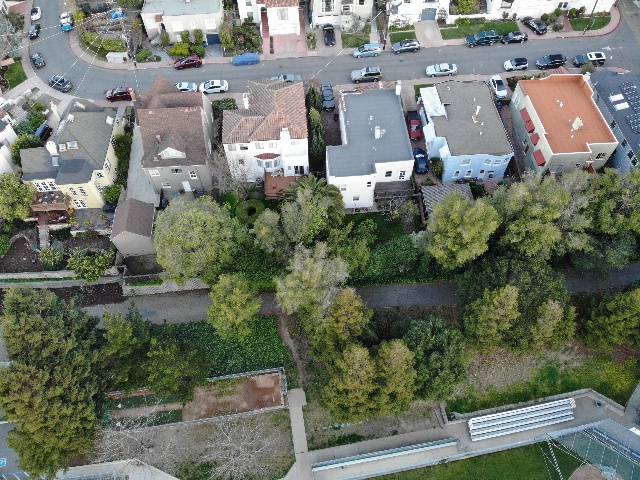}
        \caption{Query images captured at pitch angles from \SI{30}{\degree} to \SI{60}{\degree}.}
        \label{results_piedmont_diff_pitches}
    \end{subfigure}
    \par\bigskip\smallskip
    \begin{subfigure}[t]{\linewidth}
        \centering
        \vspace{-0.8ex}
        \includegraphics[width=.24\textwidth]{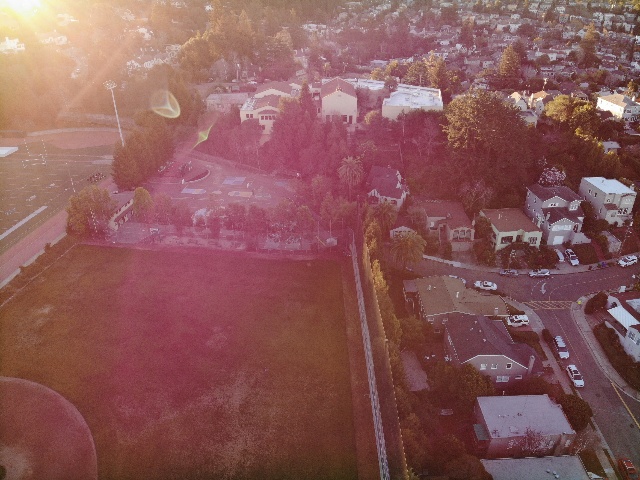}\hfill
        \includegraphics[width=.24\textwidth]{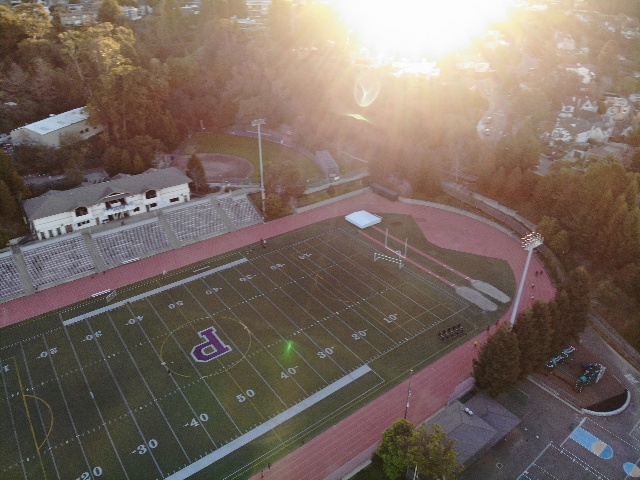}\hfill
        \includegraphics[width=.24\textwidth]{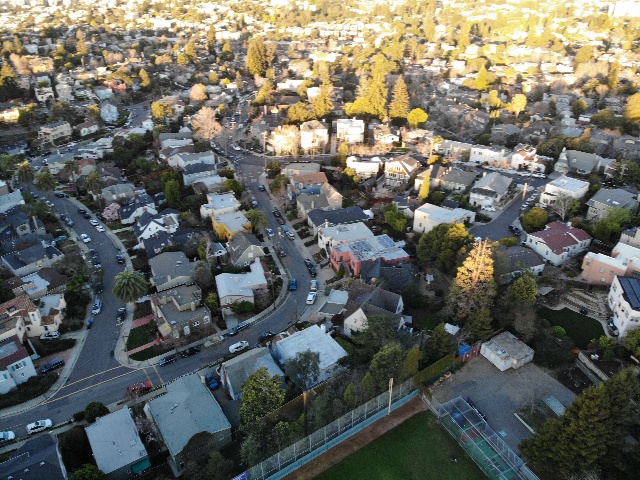}\hfill
        \includegraphics[width=.24\textwidth]{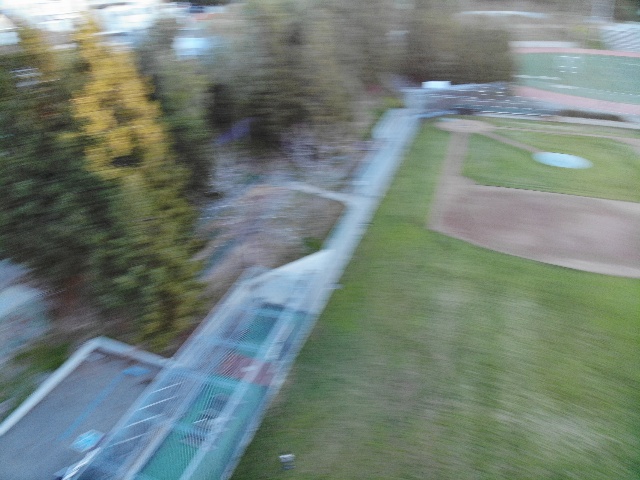}
        \caption{Sample query images collected under severe conditions, such as, from left to right, lens flare, direct sunlight, under/overexposure and motion blur, that were successfully matched and localized.}
        \label{results_piedmont_extreme_cases}
    \end{subfigure}
    \caption{Geo-localization results from the test flight at Location 2. Database was generated using satellite imagery from 2020 while query images were collected from experiments in 2021.}
    \label{results_piedmont}
\end{figure}

A similar visual presentation of the results from the second experiment at Location 2 can be found in Fig.\ \ref{results_piedmont}. This experiment was under a much more extreme condition, including variable pitch angles for the camera, image defects and lower altitudes. As shown in Fig.\ \ref{results_piedmont_altitude_dist_travelled}, the minimum localized altitude was \SI{18}{\metre} with database images rendered at a nominal altitude of \SI{60}{\metre}. Due to the additional pitch angles added to the database, shown in Fig.\ \ref{results_piedmont_diff_pitches}, query images collected at pitch angles from \SI{30}{\degree} to \SI{60}{\degree} were able to be localized. When the camera was pointed too steep down, compared with the nominal pitch angles at which database images were rendered, the visual distortion and lack of features in the view led to failures in localizing the vehicle. This happens more frequently when an extremely low altitude, such as \SI{18}{\metre} in this case, or a narrow FOV is present. From Fig.\ \ref{results_piedmont_extreme_cases}, it is noticed that the query images suffered from many environmental uncertainties. We encountered four types of interferences: lens flare, direct sunlight, high-contrast lighting and motion blur.  Despite these visual interferences, all four of the listed images were localized successfully.

In terms of the real-time performance, this stand-alone package was able to infer the pose of the vehicle at \SI{1.121}{\hertz} on the customized quadrotor shown in Fig.\ \ref{customised_quadrotor}. Quantitatively, these inferred poses are compared with the ground truth positions recorded by a GPS receiver using the root-mean-square error (RMSE) metric. The 3D and 2D RMSEs are \SI{2.816}{\metre} and \SI{2.472}{\metre} respectively along a flight path of \SI{360}{\metre} in the first experiment. The 3D RMSE includes the errors in altitude while the 2D RMSE only focuses on planar error distances. The localization was computed offline in the second experiment and the 3D and 2D RMSEs are \SI{2.418}{\metre} and \SI{2.108}{\metre} along a flight path of \SI{220}{\metre} when three initial candidates were chosen.

A more thorough benchmark of the impact on the RMSEs from different numbers of initial candidates can be found in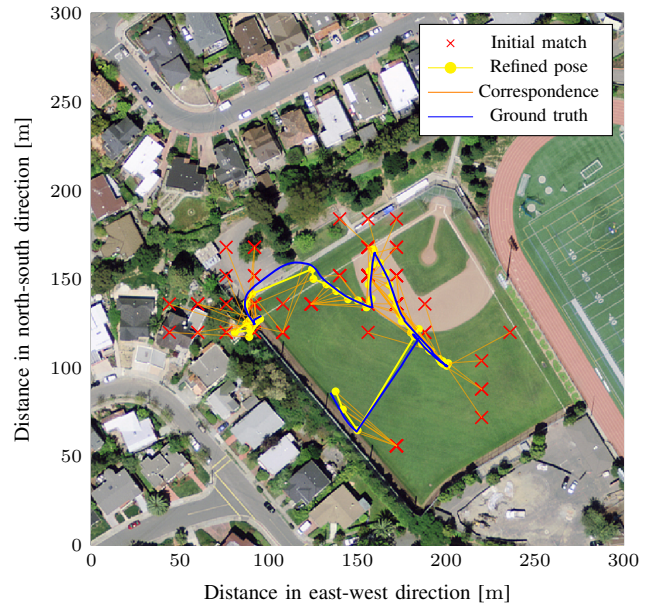
\noindent TABLE \ref{results_rmse}. In the first experiment at Location 1, a drastic reduction in the number of candidates from 50 to 1 brings negligible loss in the errors but around $30\%$ decrease in the percentage of matched images among total query images, or alternatively, the recall. At Location 2, this decrease in the percentage grows further to $45\%$. Query images with features that are more distinct and unique in the flying area have a larger chance of convergence during the global descriptor matching stage. Therefore, just a few top candidates are sufficiently probable to capture the best match in a feature-rich area, and in contrast, more candidates are required to cover the ideal match for less distinguishing images such as those with plain colors, grass and repetitive patterns.

\begin{table}
\caption{A benchmark of the impact on RMSEs and percentage match from different numbers of initial candidates at the global descriptor matching stage. We preserve these candidates to enter the local descriptor matching stage. As the number of candidates decreases, the RMSEs in both 3D and 2D cases remain approximately unchanged but the percentage of successfully localized attempts among all query images (recall) declines depending on scenarios. The choice of the number of candidates is a trade-off between accuracy and the success rate of localization.}
    \begin{subtable}{\linewidth}
    \caption{Experimental results at test Location 1.}
    \centering
    \resizebox{\linewidth}{!}{%
    \begin{tabular}{|l|c|c|c|c|c|c|}
    \hline
    \multicolumn{1}{|c|}{\multirow{2}{*}{\textbf{Metric}}} & \multicolumn{6}{c|}{\textbf{Number of initial candidates}} \\ \cline{2-7} 
    \multicolumn{1}{|c|}{} & 50             & 30    & 20    & 10    & 3              & 1     \\ \hline
    RMSE (3D) {[}m{]}      & 2.954          & 3.110 & 2.938 & 2.980 & \textbf{2.816} & 3.010 \\ \hline
    RMSE (2D) {[}m{]}      & 2.639          & 2.749 & 2.598 & 2.674 & \textbf{2.472} & 2.760 \\ \hline
    Recall {[}\%{]}      & \textbf{64.42} & 61.13 & 55.05 & 50.39 & 46.90          & 42.79 \\ \hline
    \end{tabular}%
    }
    \label{results_rmse_richmond}
    \end{subtable}\\[2.0ex]%
    \begin{subtable}{\linewidth}
    \caption{Experimental results at test Location 2.}
    \centering
    \resizebox{\linewidth}{!}{%
    \begin{tabular}{|l|c|c|c|c|c|c|}
    \hline
    \multicolumn{1}{|c|}{\multirow{2}{*}{\textbf{Metric}}} & \multicolumn{6}{c|}{\textbf{Number of initial candidates}} \\ \cline{2-7} 
    \multicolumn{1}{|c|}{} & 50             & 30             & 20    & 10    & 3     & 1     \\ \hline
    RMSE (3D) {[}m{]}      & 2.253          & \textbf{2.198} & 2.437 & 2.274 & 2.418 & 2.220 \\ \hline
    RMSE (2D) {[}m{]}      & 1.944          & \textbf{1.894} & 2.145 & 1.982 & 2.108 & 1.912 \\ \hline
    Recall {[}\%{]}      & \textbf{60.22} & 56.91          & 52.49 & 45.30 & 38.67 & 33.15 \\ \hline
    \end{tabular}%
    }
    \label{results_rmse_piedmont}
    \end{subtable}
\label{results_rmse}
\vspace{-2ex}
\end{table}

\section{Conclusions and future work}
\label{sec:conclusions}
In this paper, we proposed an image-based real-time geo-localization pipeline that features a reliable solution to modern localization needs on UAVs flying in GPS denied environments. The stand-alone pipeline executes on an embedded system at a high frequency and is able to accommodate to downward-tilted cameras. We show the practicality of our method by requiring no desired flight paths for preparing the map database and no prior knowledge of initial positions for online localization.

Some limitations and extensions of the pipeline are discussed next. Although the potential areas of search over the database can be confined according to the current position of the vehicle, the detection of outliers is still of vital importance when the very first global position is found. The current RANSAC approach applied during the P$n$P phase indiscriminately tests over the features extracted from the images. A content-aware inspector layer may be added right after the global descriptor matching stage to further verify the validity of the matches because the descriptors can be similar even between completely different images. Additionally, the refinement threshold is assigned empirically and it may also reject correct estimates. Therefore, extensive research is still necessary to further enhance the robustness by improving the outlier rejection.

\section*{Acknowledgment}
Research was sponsored by the Army Research Laboratory and was accomplished under Cooperative Agreement Number W911NF-20-2-0105. The views and conclusions contained in this document are those of the authors and should not be interpreted as representing the official policies, either expressed or implied, of the Army Research Laboratory or the U.S. Government. The U.S. Government is authorized to reproduce and distribute reprints for Government purposes notwithstanding any copyright notation herein.

\balance

\typeout{} 
{
\bibliographystyle{IEEEtran}
\bibliography{references.bib}
}
\end{document}